\documentclass[letterpaper, 10 pt, conference]{ieeeconf}  

\IEEEoverridecommandlockouts                              
\usepackage{graphicx} 
\usepackage{mathptmx} 
\usepackage{amsmath} 
\usepackage{amssymb}  
\usepackage{algorithm}
\usepackage{algpseudocode}
\usepackage{cite}
\usepackage{tikz}
\usepackage{refstyle}
\usepackage{multicol}
\usepackage{caption}
\usepackage{subcaption}
\usepackage{amssymb}
\usepackage{array}
\usepackage{color,soul}
\usepackage{breqn}
\usepackage{multirow}
\usepackage[mathscr]{euscript}
\usepackage[colorlinks=true,linkcolor=black,anchorcolor=black,citecolor=black,filecolor=black,menucolor=black,runcolor=black,urlcolor=black]{hyperref}
\usepackage{paralist}

\DeclareMathOperator*{\argmax}{arg\,max}

\title{\LARGE \bf
Learning with Training Wheels: Speeding up Training with a Simple Controller for Deep Reinforcement Learning
}
\author{Linhai Xie$^{1}$, Sen Wang$^{2}$, Stefano Rosa$^{1}$, Andrew Markham$^{1}$ and Niki Trigoni$^{1}$
\thanks{$^{1}$Xie, Rosa, Markham and Trigoni are with Department of Computer Science,
        University of Oxford, Oxford OX1 3QD, United Kingdom
        {\tt\small \{firstname.lastname\} @cs.ox.ac.uk}}%
\thanks{$^{2}$Wang is with School of Engineering and Physical Sciences, Heriot-Watt University, Edinburgh EH14 4AS, United Kingdom
        {\tt\small s.wang@hw.ac.uk}}%
}

\makeatletter
\def\BState{\State\hskip-\ALG@thistlm}
\makeatother

\begin{document}

\maketitle
\thispagestyle{empty}
\pagestyle{empty}

\begin{abstract}
Deep Reinforcement Learning (DRL) has been applied successfully to many robotic applications. However, the large number of trials needed for training is a key issue. Most of existing techniques developed to improve training efficiency (e.g. imitation) target on general tasks rather than being tailored for robot applications, which have their specific context to benefit from. We propose a novel framework, Assisted Reinforcement Learning, where a classical controller (e.g. a PID controller) is used as an alternative, switchable policy to speed up training of DRL for local planning and navigation problems. The core idea is that the simple control law allows the robot to rapidly learn sensible primitives, like driving in a straight line, instead of random exploration. As the actor network becomes more advanced, it can then take over to perform more complex actions, like obstacle avoidance. Eventually, the simple controller can be discarded entirely. We show that not only does this technique train faster, it also is less sensitive to the structure of the DRL network and consistently outperforms a standard Deep Deterministic Policy Gradient network. We demonstrate the results in both simulation and real-world experiments.
\end{abstract}

\section{Introduction}

Deep Reinforcement Learning (DRL) has been shown to be able to master complex games, even with high-dimensional input such as video games \cite{mnih2015human}. However, there are many additional difficulties to conquer when applying it to robot tasks. Among them, improving training efficiency is a realistic and urgent demand since a long-term training phase is almost impossible to be conducted in the real world. 

In the machine learning community, researchers mostly focus on algorithmic techniques for accelerating the training of DRL, such as parallel training \cite{mnih2016asynchronous} and data efficiency \cite{schaul2015prioritized}. However, these algorithms do not consider the context of a specific task, which can be valuable for training. Many robotic problems, for example, do have existing solutions which could benefit the training of DRL. Although these solutions may not be optimal, they still outperform a random exploration policy in most cases. How to fully exploit and benefit from prior approaches and tightly combine them with DRL to accelerate training is an important, yet open, topic. 

Autonomous navigation, one of the most fundamental capabilities in robotics, is a canonical scenario where this topic can be investigated.  Teaching a robot to swiftly navigate towards a target in an unknown world, whilst avoiding obstacles, requires a huge number of trials (e.g. in the order of millions) to learn a good policy. This clearly is impractical to perform in the real-world. Instead, we can exploit the close correspondence between a simulator and the real-world, to transfer the learned policy. This is especially the case when using laser range finders \cite{tai2017virtual} or depth images \cite{xie2017towards}.

\begin{figure}
  \centering
  \includegraphics[width=.7\linewidth]{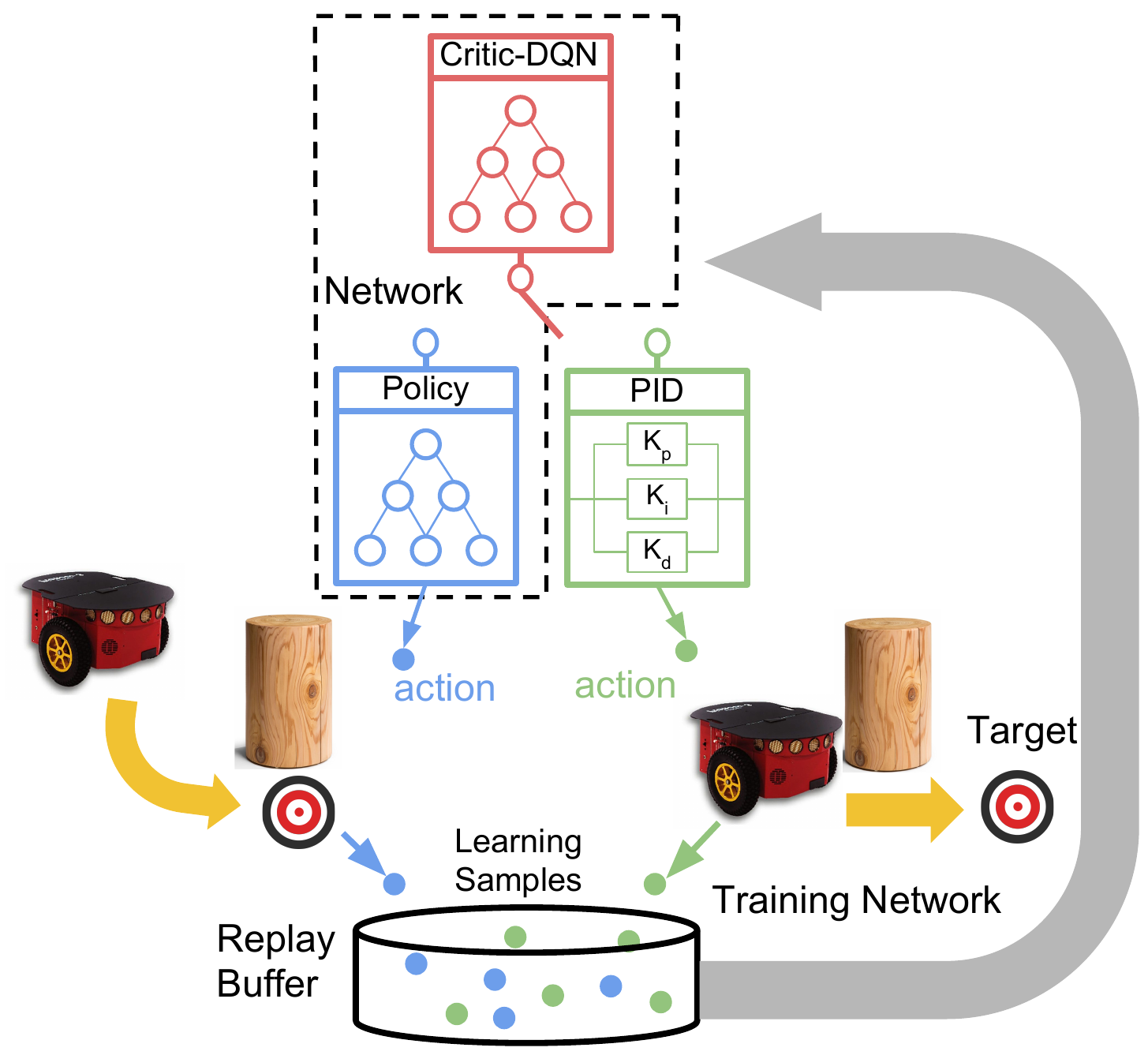}
  \label{fig:front-page}
  \caption{\small A deep neural network is trained with an actor-critic Reinforcement Learning approach to learn local planning for robot navigation. The critic-DQN network assesses both the performance of the external controller, e.g. a PID controller, and the policy network, selecting actions from a better one according to the situation. All the resulting learning samples are stored in the replay buffer. Therefore, the policy network can improve itself either by imitating the external controller or by examining its own policy.\normalsize}
\end{figure}

One strategy that allows reinforcement learning to benefit from an existing controller is to generate training labels for input states by using self-supervised or semi-supervised learning\cite{pfeiffer2017perception,yang2017obstacle}. Another approach is to generate a few demonstration samples with high performance and ask the networks to imitate them, i.e. imitation learning\cite{duan2017one}. However, in the autonomous navigation problem domain, a good controller itself is difficult to design. Instead, we consider using a simple control law e.g. a proportional (P) controller. As this will be unable to navigate past obstacles, we cannot use such a naive approach to record a demonstration trace. However, it will obtain a higher reward on average compared with a completely random strategy which is a common policy exploration approach. Our intuition is to use this controller like \emph{training wheels} on a bicycle - they prevent a novice from falling off in the beginning, by making the control problem easier, but once the rider has mastered how to balance, they can be safely removed.

In this paper, we present a novel framework, called \emph{Assisted Reinforcement Learning}, which is able to significantly accelerate and improve the training of DRL by incorporating an external controller. Built upon this framework and Deep Deterministic Policy Gradient (DDPG) \cite{lillicrap2015continuous}, we propose the \emph{Assisted Deep Deterministic Policy Gradient (AsDDPG)} algorithm. Our contributions are summarized as follows:

\begin{itemize}
\item We propose a novel actor-critic algorithm that can seamlessly incorporate an external controller to assist DRL and eventually work independently from the controller.
\item Training of DDPG is significantly accelerated and stabilized for robot navigation with the AsDDPG strategy.
\item Autonomous navigation is achieved in the framework of DRL with fast training, showing promising results in challenging environments.
\end{itemize}

The rest of this paper is organized as follows. Related work is reviewed in Section \ref{sec:RelatedWork}. The background and the proposed AsDDPG algorithm are described in Sections \ref{sec:background} and \ref{sec:algorithm}, respectively. Section \ref{sec:ExpResults} presents experimental results, followed by conclusions in Section \ref{sec:Conclusion}. Note that the implementation is available at \url{https://github.com/xie9187/AsDDPG}.

\section{Related Work}\label{sec:RelatedWork} 

Deep Learning (DL) has garnered an intense amount of attention in the robotics community due to its performance in a number of different and complex tasks, e.g. localization\cite{wang2017deepvo,Clark_AAAI17}, navigation\cite{pfeiffer2017perception} and manipulation\cite{levine2016learning}.

\subsection{Supervised Deep Learning in Robot Navigation}

Robot navigation is a well studied problem and a large amount of work has been developed to tackle issues of autonomous navigation \cite{oriolo1995line,kim1999symbolic}. Recently, several supervised and self-supervised DL approaches have been applied to navigation \cite{giusti2016machine,pfeiffer2016perception} and its sub-problems e.g. obstacle avoidance \cite{yang2017obstacle}. However, limitations prevent these approaches from being widely used in a real robotic setting. For example, a massive manually labeled dataset is required for the training of the supervised learning approaches. Although this can be mitigated to an extent by resorting to self-supervised learning methods, their performance is largely bounded by the strategy generating training labels.

\subsection{Deep Reinforcement Learning in Robot Navigation}

Different from previous supervised learning methods, DRL based approaches learn from a large number of trials and corresponding rewards instead of labeled data. For instance, Zhu et al.\cite{gandhi2017learning} train a network which can steer a monocular-based robot to find an image as a target by only giving a large reward when the target is found.

However, because of the excessive number of trials required to learn a good policy, training in a simulator is more suitable than experiences derived from the real world. In \cite{sadeghi2016cad}, a network learns a controller for a flying robot to avoid obstacles through monocular images. The learned policy is transferred from simulation to reality by frequently changing the rendering settings of the simulator to bridge the gap between the images from the simulator and real-world. %
When utilizing laser scans instead of images as input, DRL models trained in the simulator can be directly applied in real world \cite{tai2017virtual}. As an alternative approach, in~\cite{xie2017towards} the differences between RGB images in simulation and reality are mitigated by first transferring the RGB images to depth images.

\subsection{Accelerating Training}

In this work, we pay more attention to improving the training efficiency, to reduce the number of trials required. \cite{zhang2016deep} achieves this by transferring knowledge learned for navigation through successor features but is limited to operate within similar scenarios. In \cite{mnih2016asynchronous}, training is sped up by executing multiple threads in parallel, but this can result in more computing overhead especially when the simulator is computationally heavy. Gu et al.~\cite{gu2016continuous} propose an additional network that is trained to learn the model of the environment. This can be used for generating more training data, accelerating the training procedure with these additional synthetic samples. Zhang et al.~\cite{zhang2016learning} apply a traditional model predictive controller (MPC) to assist the training of the network controlling a drone. However, the controller is utilized to initialize the network in a supervised style. The point at which to transfer to Reinforcement Learning policy needs to be manually decided. Vecerik et al.~\cite{vevcerik2017leveraging} exploit human demonstrations to accelerate the training of DDPG in manipulation, but this suffers from two limitations. Firstly, they manually inject demonstration samples into the replay buffer~\cite{schaul2015prioritized}. Secondly, the human demonstrations needed take significant time and effort to collect.

Our work is also related to Hierarchical Reinforcement Learning as described in \cite{tessler2017deep} which switches among different Neural Skill Networks to accelerate learning in a lifelong term. But unlike \cite{tessler2017deep} which keeps using the switch all the time, our AsDDPG can finally discard the external guidance and only utilize a simple policy network to complete the task.

\section{Background}\label{sec:background}

In this work, we focus on training a network as a local planner to deal with the robot navigation problem, that is, it is designed to drive the robot to a nearby target without colliding with obstacles. Our proposed approach is based on two DRL methods: Deep Q networks (DQN)~\cite{mnih2015human} and DDPG~\cite{lillicrap2015continuous}. They will be briefly introduced in this section after we outline the generic problem of robotic navigation.

\begin{figure*}
  \centering
  \includegraphics[width=0.7\linewidth]{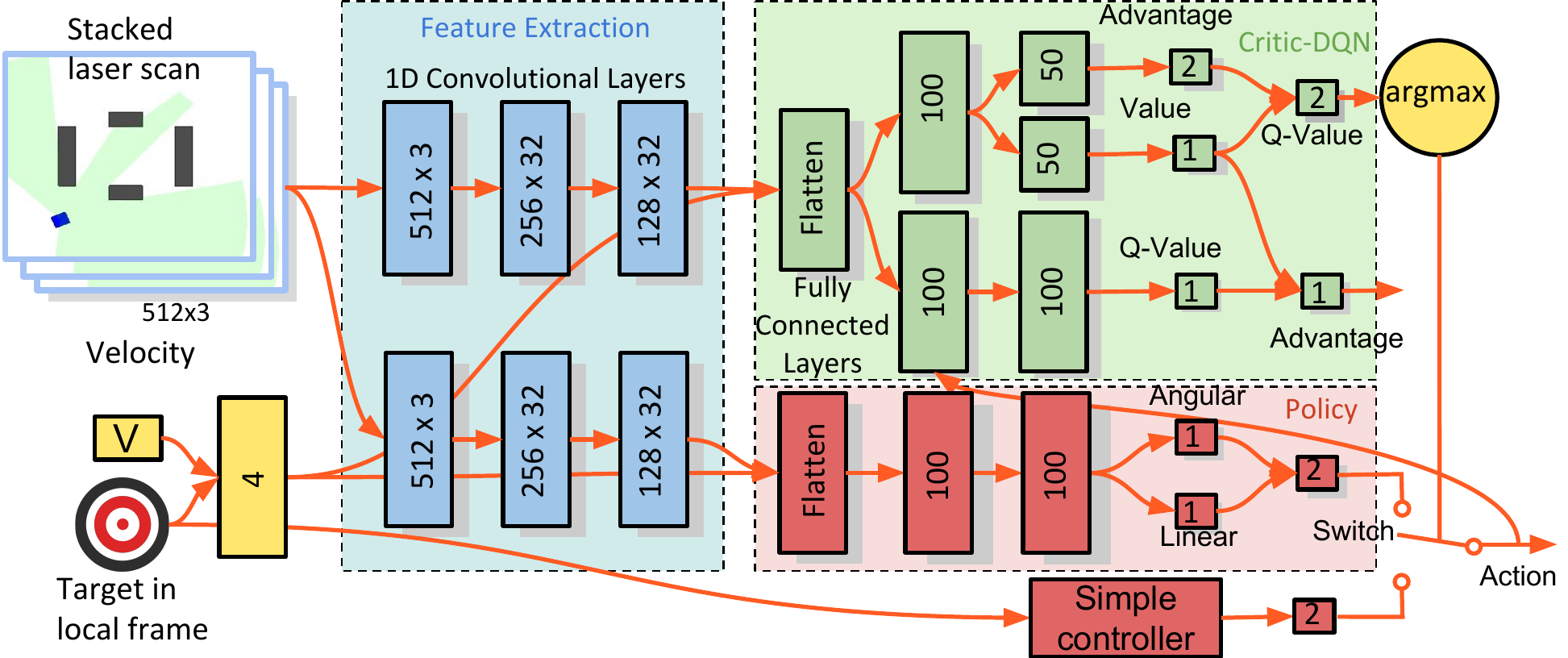}
  \caption{\small Network architecture. Network layers are demonstrated by the rectangles. Orange arrows indicate the connectivity between network layers and some other components, e.g. input state and the output of the simple controller. The final action is selected based on the Q value predicted by critic-DQN.}
  \label{fig:network}
\end{figure*}

\subsection{Problem Formulation}

We can consider the local navigation problem as a decision making process where the robot is required to avoid obstacles and reach a target position. At time $t \in [0,T]$ the robot takes an action $a_t \in \mathscr{A}$ according to the input $x_t$. The input contains a view of the world e.g. a stack of laser scans, the current speed of robot, and the target position with respect to the robot's local frame. We assume that the robot can localize itself in the global coordinate frame with a map which enables the calculation of the target position in local frame. After executing the action, the robot receives a reward $r_t$ given by the environment according to the reward function and then transits to the next observation $x_{t+1}$. The goal of this decision making procedure is to reach a maximum discounted accumulative future reward $R_t = \sum_{\tau=t}^{T}\gamma^{\tau-t}r_\tau$, where $\gamma$ is the discount factor. 

\subsection{Deep Q Network (DQN)}

A DQN is a RL algorithm based on Q learning and deep neural networks. It only estimates the value of a state-action pair $(x_t, a_t)$, which is termed the Q-value for all states. Given the policy $a_t = \pi(x_t)$, it can be defined as follows 
\begin{equation}
  \begin{aligned}
    Q^\pi(x_t,a_t) &= \mathbb{E}[R_t|x_t,a_t,\pi],\\
  \end{aligned}
\end{equation}
which can be calculated with the Bellman equation
\begin{equation*}
	Q^\pi(x_t,a_t) = \mathbb{E}[r_t+\gamma\mathbb{E}[Q^\pi(x_{t+1},a_{t+1})|x_t,a_t,\pi].
\end{equation*}
By choosing the optimal action each time where $Q^*(x_t,a_t)=max_\pi\mathbb{E}[R_t|x_t,a_t,\pi]$, we can have the optimal Q-value function
\begin{equation}
Q^*(x_t,a_t)=\mathbb{E}_{x_{t+1}}[r+\gamma \max\limits_{a_{t+1}}Q^*(x_{t+1},a_{t+1})|x_t,a_t],
\label{eq:optimal q}
\end{equation}
which shows that by adding the discounted optimal Q-value at time $t+1$ with the current reward, the optimal Q-value at time $t$ can be approximated rather than computed directly over a large state space. In summary, DQN utilizes a deep neural network (parameterized by $\theta^Q$) to estimate the Q-value through Q-learning.

\subsection{Deep Deterministic Policy Gradient (DDPG)}

Similar to DQN, DDPG~\cite{lillicrap2015continuous} also estimates the Q-value for each state-action pair with a critic network which is parameterized by $\theta ^Q$. But it also utilizes an actor network (parameterized by $\theta ^\pi$) to estimate optimal actions directly, which will be assessed by the critic network. Such an actor-critic architecture makes it suitable to work in a continuous action domain which is difficult for DQN and is ideal for controlling robots. DQN applies a greedy policy where we need to maximize the Q-value w.r.t. the actions. Thus, if the action is continuous, it will be computationally expensive. This is one of the reasons why DDPG is appealing in robotics as most of the robotic tasks stay in a continuous action domain.

The training for critic network is almost the same as DQN, but the actor network is updated with policy gradient, defined through the chain rule as follows:
\begin{equation}
\begin{split}
\triangledown _{\theta^{\pi}}\pi \approx \mathbb{E}[\triangledown _{\theta^{\pi}}Q(x, a|\theta^Q)|_{x=x_t, a=\pi(x_t|\theta^\pi)}]= \\
\mathbb{E}[\triangledown _aQ(x,a|\theta^Q)|_{x=x_t,a=\pi(x_t)}\triangledown _{\theta ^\pi} \pi(x|\theta ^\pi)|_{x=x_t}].
\label{eq:policy gradient}
\end{split}
\end{equation}
It indicates that the policy gradient can be obtained by multiplying two partial derivatives. One is the derivative of Q-value $Q(x,a|\theta^Q)$ obtained from the critic network w.r.t. the output action $a=\pi(x_t|\theta^\pi)$ by actor network and the other is action $a$ w.r.t. the parameters of actor network $\theta^\pi$.

\section{Assisted Deep Reinforcement Learning}\label{sec:algorithm}

The main insight behind our framework is to provide a simple controller to assist the network in policy exploration, accelerating and stabilizing the training procedure. We in particular focus on DDPG, and hence term our approach \emph{Assisted} DPPG, or AsDDPG for short. The intuition is simple: a naive control law will outperform a random strategy for simple tasks e.g. driving in a straight line.   

However, instead of simply treating this controller as an independent exploration method like $\epsilon$-greedy, we combine the critic network with a DQN to automatically judge which policy it should use to maximize the reward. Essentially, the augmented critic network controls a switch which determines whether the robot follows the controller's suggested actions or the learned policy. This can avoid manually tuning parameters to decide when and how to use such an external controller, potentially deriving an optimal strategy. Furthermore, this external controller does not need to solely cope with the whole task. Instead, it is only used to outperform sub-optimal policies (e.g. random actions). 

Since AsDDPG is an off-policy learning method where the network learns from a replay buffer, regardless of the current policy, the actor network can benefit from learning samples generated by both its own recorded policy and the assisting controller. Initially, the simple controller will be chosen more frequently as the optimal policy. However, over time, the learned policy will outperform the simple controller in terms of total reward. Once training has converged, both the critic and the external controller can be discarded, and the robot simply navigates based on the learned policy.


\subsection{Network Architecture}

To bring the previously discussed intuition into practice, we design a novel network architecture shown in Fig.~\ref{fig:network}. It includes three parts, namely feature extraction (blue), policy network and assistive controller (red), and the augmented critic network (green). 

The first part of the system is the 1-D convolutional layers which are utilized to extract features from the stacked dense laser scans. The activations applied are ReLU. We find these convolutional layers to be typically important for the policy to reach both a good performance for obstacle avoidance and an acceptable generalization ability in the real world. In~\cite{tai2017virtual}, the author only uses a sparse laser scan (10 beams of a scan) which enables good generalization in different scenarios. However, it is difficult for the robot to avoid small obstacles smoothly. Intuitively, decimating a high-fidelity observation loses information and is not ideal. Thus, we prefer to keep the dense laser scans as input and instead apply 1-D convolutional layers to learn efficient features for our task. Stacking inputs across multiple timestamps also provides more information on the environment. 

The second part is the policy network with fully-connected layers, estimating the optimal linear and angular speeds for the robot based only on features extracted from the input state (e.g., laser scans, current speed and target position in local frame). Note that the activations for these two outputs are sigmoid and $\tanh$, respectively. The external controller also generates a control signal (policy) based on the error signals between current and target positions.

Finally, the Critic-DQN constructed with fully-connected layers is the third part. It has two branches: one is the critic branch where the action is concatenated into the second layer; the other is DQN branch where we apply dueling~\cite{DBLP:journals/corr/WangFL15} and double network architecture~\cite{van2016deep} to speed up the training and avoid overestimation. Note that there is no nonlinear activation for its output layers. We discuss the critic in more detail below.

\begin{algorithm}
\caption{AsDDPG}\label{euclid}
\begin{algorithmic}[1]
\Procedure{Training}{}
\State Initialize $A(x,a|\theta ^A)$, $Q(x,\sigma|\theta ^Q)$ and $\pi(x|\theta ^\pi)$.
\State Initialize target network $\theta ^{A'}$, $\theta ^{Q'}$ and $\theta ^{\pi'}$
\State Initialize replay buffer R and exploration noise $\epsilon$
\For{episode=1, M}
	\State Reset the environment
    \State Obtain the initial observation
    \For{step = 1, T}
     	\State Infer switching $[Q_{policy}, Q_{P}] = Q(x_t,\sigma|\theta ^Q)$
        \State $\sigma_t = argmax([Q_{policy}, Q_{P}])$
        \If{$\sigma_t==1$}
        	\State Sample policy action $a_t = \pi(x_t|\theta ^\pi) + \epsilon$
        \Else
        	\State Sample action $a_t$ from external controller
        \EndIf
    	\State Execute $a_t$ and obtain $r_t, x_{t+1}$
        \State Store transition $(x_t, a_t, r_t, x_{t+1}, \sigma_t)$ in R
        \State Sample N transitions $(x_i, a_i, r_i, x_{i+1}, \sigma_t)$ in R
        \State Optimise critic-DQN by minimising Eq.~\ref{eq:loss function}
        \State Update the policy according to Eq.~\ref{eq:policy gradient through advantage}
        \State Update the target networks
    \EndFor
\EndFor
\EndProcedure
\end{algorithmic}
\end{algorithm}

\subsection{Critic-DQN}

The two branches of critic-DQN act respectively as 
\begin{inparaenum}[1)]
\item a criticizer, evaluating the action output from the policy network and generating policy gradients, and 
\item a switch, deciding when to use the learned policies from the network or the external controller.
\end{inparaenum}

The critic branch is similar to the original DDPG. However, it estimates the advantage $A^\pi(x,a)$ from Q-value $Q^\pi(x,a)$ for each state-action pair. This is leveraged by a dueling network in DQN branch where the value $V^\pi(x)$ of each state will also be learned. With the definition of advantage, it can be simply calculated as $A^\pi(x,a)=Q^\pi(x,a)-V^\pi(x)\approx Q^\pi(x,\sigma)-V^\pi(x)$. Note that the action considered by critic-DQN branch is the switching action $\sigma$ rather than $a$. According to~\cite{sutton2000policy}, compared with Q-value, estimating advantage largely reduces the variance and is essential for fast learning especially for approaches based on policy gradient. The entire critic branch is denoted as $A(x,a|\theta ^A)$ while the part before estimating the advantage is denoted as $Q^A(x,a|\theta ^A)$. 

The DQN branch estimates and compares the Q-values for either using the policy from the actor network or applying actions from the external controller based on the state inputs. It greedily switches to the one with a higher Q-value estimate. The DQN branch is denoted as $Q(x,\sigma|\theta ^Q)$.

\subsection{Gradients for Training}

The learning samples can be defined as a tuple $(x_t, a_t, r_t, x_{t+1}, \sigma _t)$, where $\sigma _t$ is a binary variable, indicating the switching action. Among these variables, $a_t$ will only affect the update for the critic branch while $\sigma _t$ is only for the DQN branch.

In the critic-DQN, both branches optimize their network parameters through boot-strapping. This means they learn from the temporal-difference (TD) error of the Q-value estimation with Eq.~\ref{eq:optimal q}. More specifically, weights are optimized based on the following loss function $L$:
\begin{equation}
  \begin{aligned}
    y_i^A&=r_i+\gamma Q^{A'}(x_{i+1},\pi '(x_{i+1}|\theta^{\pi'})|\theta^{A'}) \\
    y_i^Q&=r_i+\gamma Q'(x_{i+1},\argmax_{\sigma}Q(x_{t+1}, \sigma_{t+1}|\theta^{Q}))|\theta^{Q'}) \\
    L&=\dfrac{1}{N}(\sum_i(y^A-Q^A(x_i,a_i|\theta^A))^2+(y^Q-Q(x_i,\sigma_i|\theta^Q))^2).
    \label{eq:loss function}
  \end{aligned}
\end{equation}
where $Q^{A'}$, $Q'$ are target networks for the two branches and $i$ is the indices of samples in the batch. Note that the critic branch is updated through its Q value estimation instead of the final output advantage which is used for generating policy gradient.

For the actor network, its weights are adjusted with policy gradients which are defined in Eq.~\ref{eq:policy gradient}. It requires the critic branch to first compute the gradients of its advantage output $A^\pi(x,a)$ w.r.t. the action $a$. This is then transferred to the actor network to calculate the gradients w.r.t. the network parameters $\theta ^\pi$. It can be derived as follows:
\begin{equation}
\begin{split}
\triangledown _{\theta^{\pi}}\pi \approx \mathbb{E}[\triangledown _{\theta^{\pi}}A(x, a|\theta^A)|_{x=x_t, a=\pi(x_t|\theta^\pi)}]= \\
\mathbb{E}[\triangledown _a A(x,a|\theta^A)|_{x=x_t,a=\pi(x_t)}\triangledown _{\theta ^\pi} \pi(x|\theta ^\pi)|_{x=x_t}].
\label{eq:policy gradient through advantage}
\end{split}
\end{equation}

Algorithm~\ref{euclid} outlines the entire training process of AsDDPG.

\section{Experiments}\label{sec:ExpResults}

\begin{figure}[t]
  \centering
  \begin{subfigure}{.25\linewidth}
      \centering
      \includegraphics[width=\linewidth]{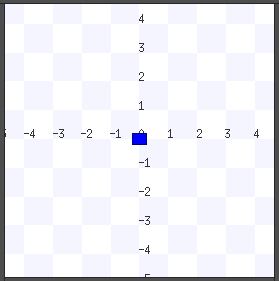}
      \caption{Empty}
      \label{fig:stage empty}
  \end{subfigure}
  \begin{subfigure}{.28\linewidth}
      \centering
      \includegraphics[width=\linewidth]{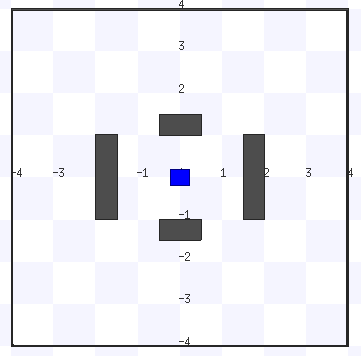}
      \caption{Simple}
      \label{fig:stage simple}
  \end{subfigure}
  \begin{subfigure}{.28\linewidth}
      \centering
      \includegraphics[width=\linewidth]{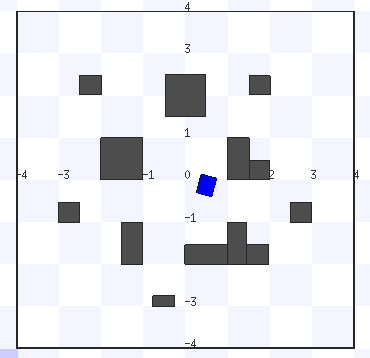}
      \caption{Complex}
      \label{fig:stage complex}
  \end{subfigure}
  \caption{\small The three Stage simulation worlds used for training. The gray rectangles are obstacles while the blue one is the robot. The target position is randomly generated for each episode. \normalsize}
  \label{fig:stage}
\end{figure}

Several experiments are conducted to evaluate the performance of the proposed AsDDPG against the original DDPG for the robot navigation problem. We train the networks in the Stage and Gazebo simulators and test the learned policy in a real world scenario.

\begin{figure}[t]
  \centering
  \includegraphics[width=0.8\linewidth]{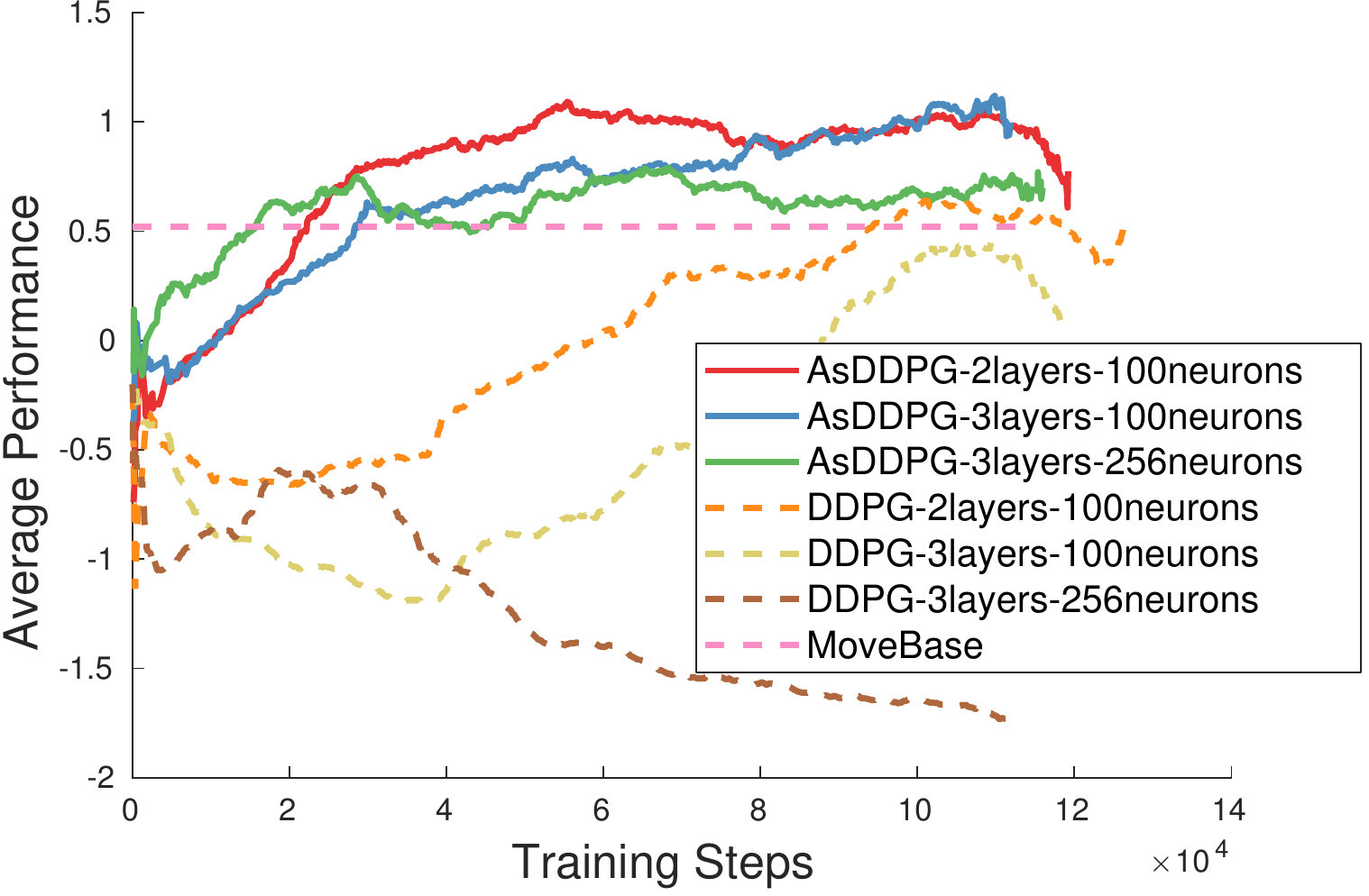}
  \caption{\small Smoothed learning curves with different network hyper-parameters. It illustrates the average performance of the network at each training step. Note that MoveBase is selected as the baseline approach whose average performance reaches at about 0.5. \normalsize}
  \label{fig:learning curves parameter}
\end{figure}

\begin{figure}[t]
\centering
\begin{subfigure}{.5\linewidth}
  \centering
  \includegraphics[width=.9\linewidth]{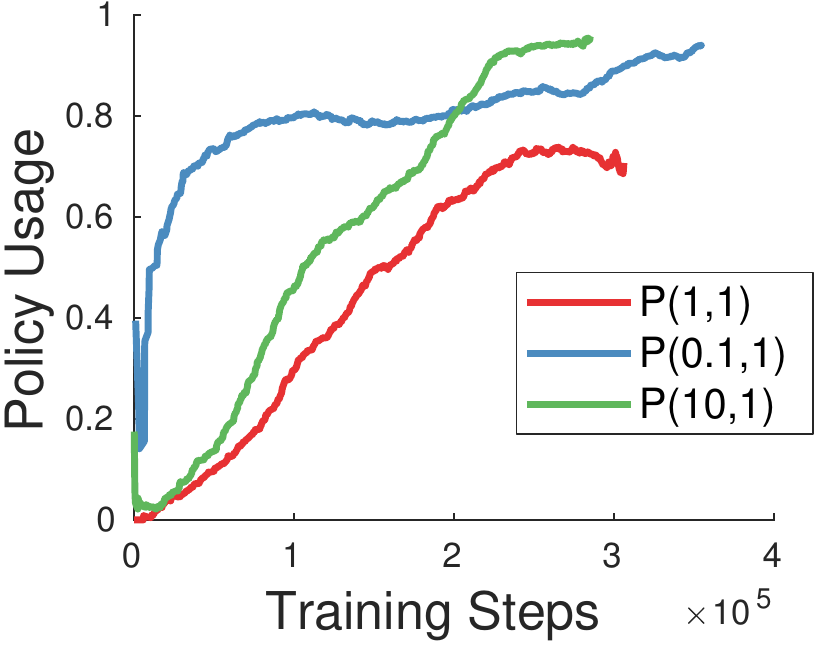}
  \caption{Empty world}
\end{subfigure}%
\begin{subfigure}{.5\linewidth}
  \centering
  \includegraphics[width=.9\linewidth]{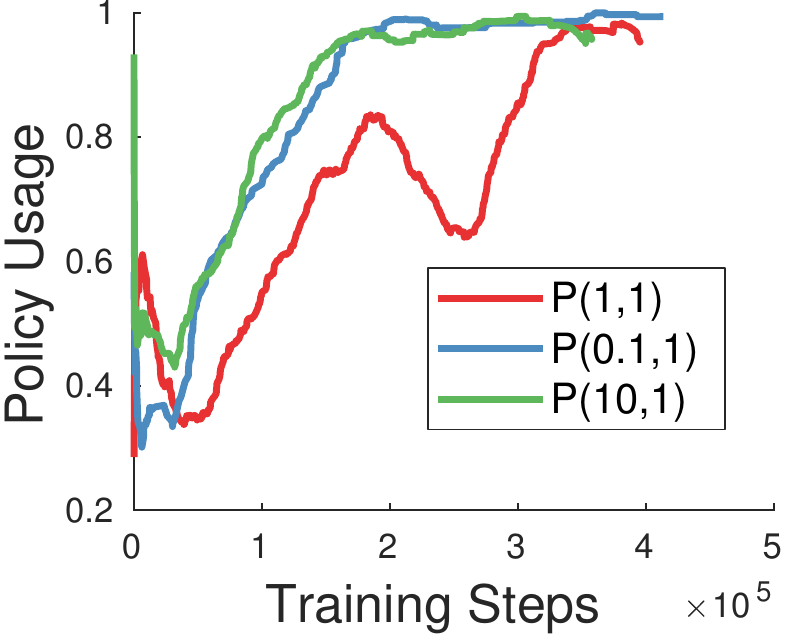}
  \caption{Simple world}
\end{subfigure}
\caption{\small The usage of the policy network within an episode through the entire training procedure respectively in two different simulation worlds.\normalsize}
\label{fig:different p pidrate}
\end{figure}

For the training in the Stage simulator, there are three environments as shown in Fig.\ref{fig:stage}. Both DDPG and AsDDPG are trained with the same reward function. Specifically, it contains a sparse part $R_{reach}$ and $R_{crash}$, where the robot obtains a large positive reward for reaching the goal and a large negative reward for colliding with an obstacle, and a dense part as
\[
    r_t = 
\begin{cases}
    R_{crash}, & \text{if crashes}\\
    R_{reach}, & \text{if reaches the goal}\\
    \gamma_p((d_{t-1} - d_t)\triangle t - C), & \text{otherwise}
\end{cases}
\]
where $d_{t-1}$ and $d_t$ indicate the distance between robot and target at two consecutive time stamps, $\triangle t$ represents the time for each step, $C$ is a constant used as time penalty. and $\gamma_p$ is a discount factor.

The default value of $\gamma_p$ is $1$ for using policy network. But its value can be set to a smaller number (e.g. $0.5$) for penalizing the usage of external controller when the dense reward is positive. This serves for two purposes: i) experiencing with the policy network more frequently and ii) learning faster on how to work independently of the external controller. Theoretically, the network can learn to gradually diminish the usage of the external controller since there is always a better policy instead of utilizing the external controller in terms of reward.

\begin{figure*}[h]
\centering
\begin{subfigure}{.4\linewidth}
  \centering
  \includegraphics[width=\linewidth]{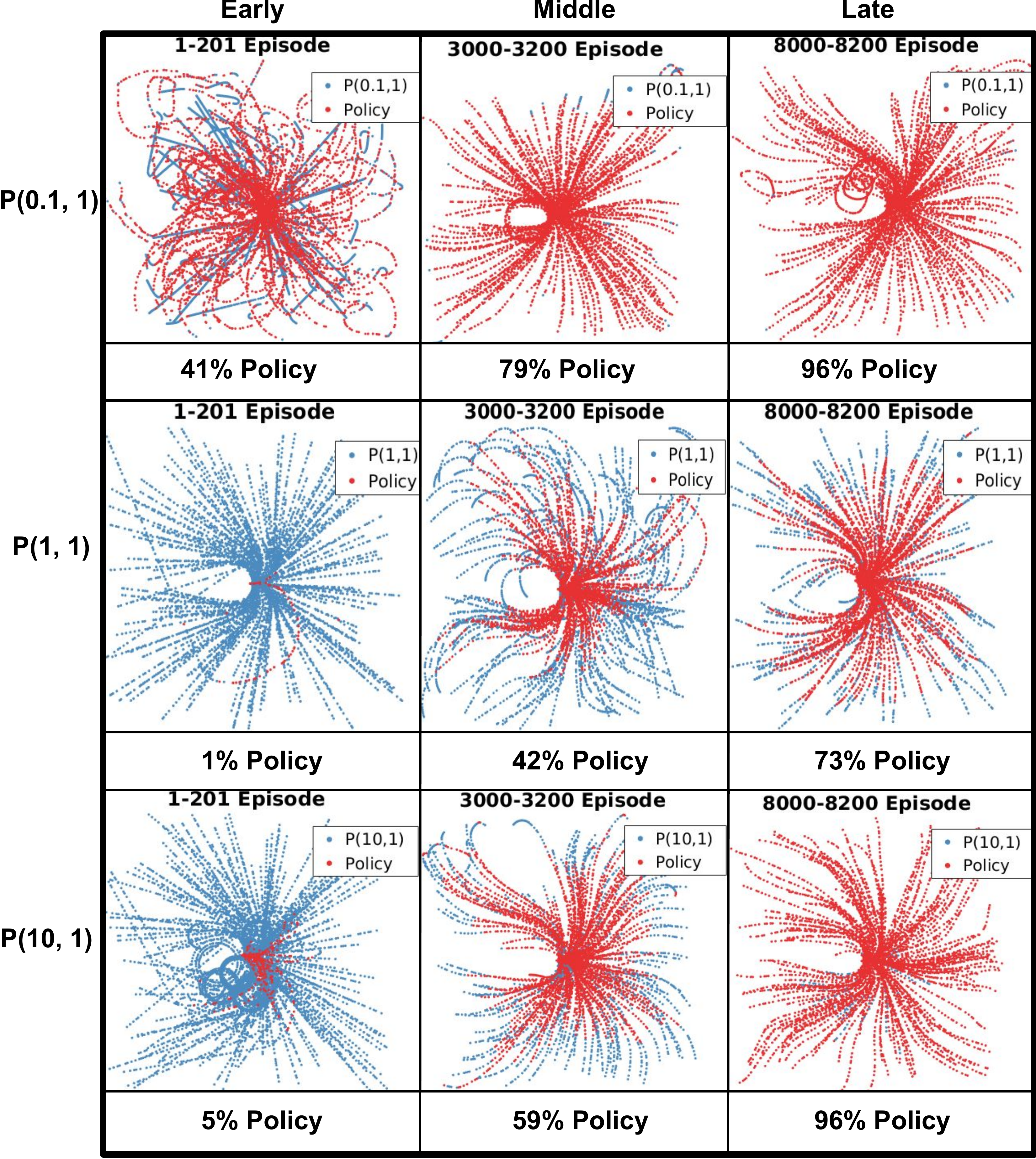}
  \caption{Empty world}
  \label{fig:different p traj switch empty}
\end{subfigure}
  \qquad
\begin{subfigure}{.4\linewidth}
  \centering
  \includegraphics[width=\linewidth]{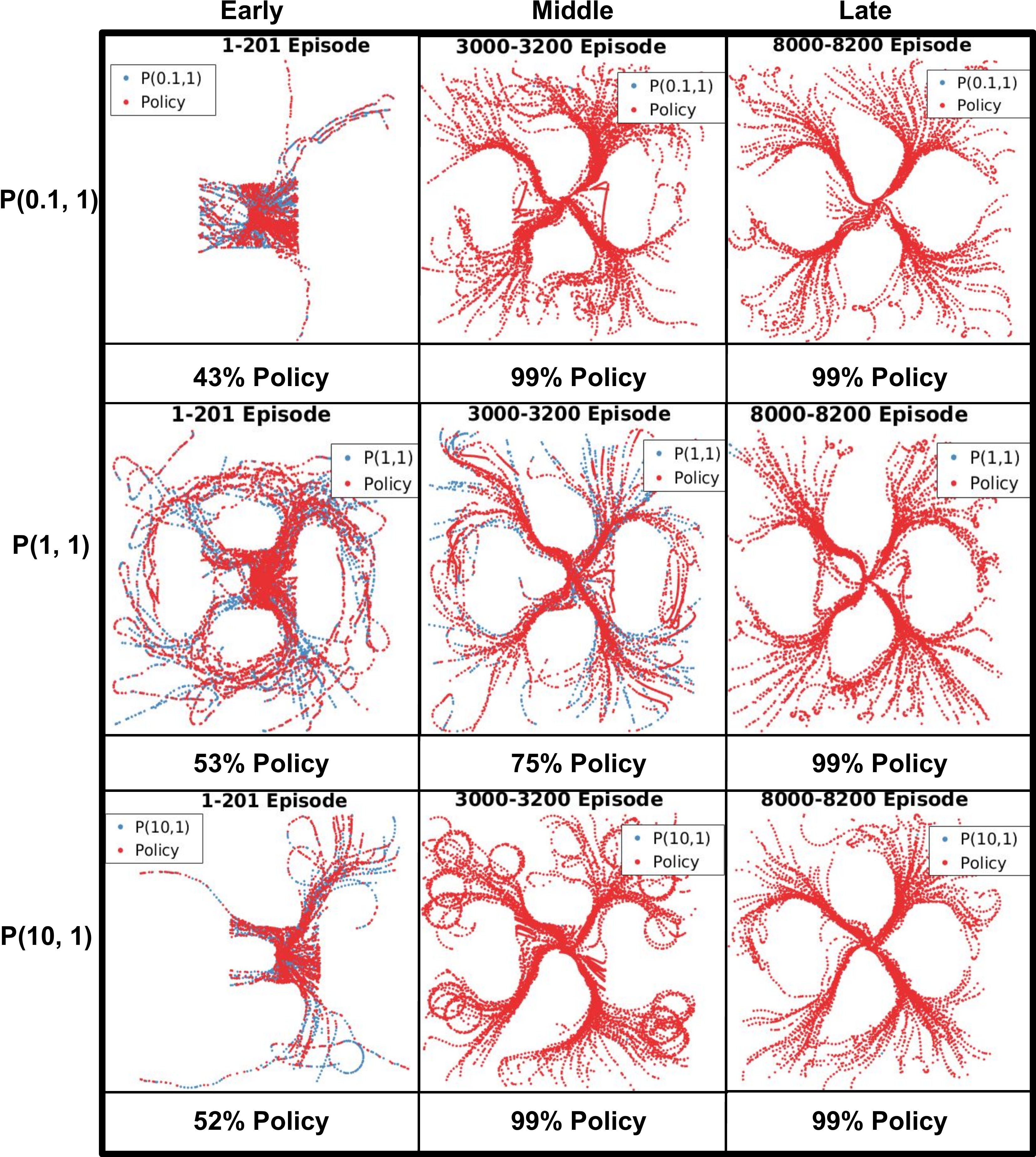}
  \caption{Simple world}
  \label{fig:different p traj switch obs}
\end{subfigure}
\caption{\small The policy learned by the robot with various proportional controller at different stages of the training. Each image illustrates the trajectories of the robot as well as the switching results between the external controller and policy network among 200 episodes, respectively at the early, middle and late phases of the training procedure.\normalsize}
\end{figure*}


The reward function above does not necessarily provide an objective performance metric since it is designed to alleviate the training difficulty. Hence, in this experiment we use a navigation task metric based on the time taken to reach the goal and whether the robot reaches the goal. More specifically, each step gives a time penalty of $-0.01$ and reaching the goal gives a positive reward of $2$.

\subsection{Speeding up Training Procedure with Various Hyper-parameters}
A known problem of DDPG is the high sensitivity to network hyper-parameters. Manually tuning hyper-parameters to make DDPG converge is very time-consuming and is something that ideally could be avoided. Therefore, in this experiment, we examine networks with three distinct settings, including two or three fully connected layers with 100 neurons in each layer or three layers with 256 neurons. 

The resulting learning curves over more than 100k training steps for the different hyper-parameters are shown in Fig.~\ref{fig:learning curves parameter}. It demonstrates that AsDDPG outperforms DDPG in terms of training efficiency, and is more stable than DDPG with different network hyper-parameters. Note that DDPG with three layers and 256 neurons per layer fails to learn a reasonable policy. Furthermore, we set MoveBase\footnote{http://wiki.ros.org/move_base} as the baseline approach to show that the learnt policy is comparable with existing deterministic approach.

\subsection{Impact of Controller Parameters}


In this experiment, several models are trained with different controller parameters in two Stage worlds (Fig~\ref{fig:stage empty} and Fig~\ref{fig:stage simple}) to investigate the sensitivity of the AsDDPG to the controller parameters, i.e., how the controller parameters affect the training. We also examine how often the critic-DQN chooses the learned policy over the external controller, studying whether it can gradually become independent and learn a good policy.


The proportional controller $P(P_l,P_r)$ is configured with two parameters. It controls linear and rotational velocity as $v = P_l\cdot x_{local}$ and $\omega = P_r\cdot y_{local}$, where $x_{local}$ and $y_{local}$ are the coordinates of the target in the robot's local coordinate frame. We investigate three settings of the controller by altering the linear gain parameter, namely $P(0.1,1)$, $P(1,1)$ and $P(10,1)$. $P(10,1)$ is the fastest controller, but suffers from overshooting of the target, requiring the robot to turn around.  $P(0.1,1)$ is the slowest controller, making very gradual changes to the robot's speed with consequent slow acceleration. $P(1,1)$ is a controller that gives good performance without serous overshoot. 

Fig.~\ref{fig:different p pidrate} shows how the ratio between the policy and the external controller chosen by the critic-DQN evolves over time during training. Firstly, it can be seen that the critic-DQN learns to sample less frequently from the controller over time, with more actions coming from the policy, although the parameters of the controller and the environment decides how fast this trend can be. One obvious phenomenon is that in both simulated worlds the network drops the slow controller $P(0.1, 1)$ rapidly since it takes a longer time to reach the goal in general and it is easy for the policy network to overperform it. However, the ratio between policy and controller for $P(1, 1)$ and $P(10, 1)$ differs in the two worlds. In Fig.~\ref{fig:different p traj switch empty} and ~\ref{fig:different p traj switch obs} this behaviour is presented in more detail with robot's trajectories at different training stages. These also show how often the critic network chooses the controller (blue trace) over the learned policy (red policy) as training progresses.

According to the results shown in Fig.~\ref{fig:different p traj switch empty}, regardless of the controller, the network first learns to apply its own policy for the first few steps, as shown by a concentration of red around the origin. This is because the total reward is heavily influenced by the initial heading. Considering the fastest controller $P(10,1)$, it can be seen that initially the robot sometimes circles around the target, due to excessive speed. This problem is solved by the policy network by choosing a better heading at the beginning, demonstrating that the network can learn when to properly use the external controller.


With $P(1, 1)$, it adopts a more accurate heading towards the target and navigates to it straightly. In the middle stage, the policy network learns a more smooth but less optimal policy in terms of time. This could be a side effect of penalizing the usage of the external controller through reward function. However, we actually find this penalty essential for stabilizing the switching strategy. Eventually, the path to the target learned by the policy network becomes more straight and optimal. Since $P(1, 1)$ is a good controller, the network does not discard it until more than 8000 training episodes.

Fig.~\ref{fig:different p traj switch obs} shows the robot trajectories when training in the simple world environment. It can be seen that the network learns some distinct behaviours. The external controller is sampled less frequently than in the empty world even at the beginning. This is reasonable since a pure PID controller cannot deal with obstacle avoidance. Moreover, the learning speed is different with different controllers. For example, the network learns to go around obstacles more easily with $P(1,1)$ than with others. With $P(10, 1)$, the network learns much slower due to the confusing guidance from the controller. Although the network drops the controller early, it retains some undesirable behaviours like hovering around the target at the middle stage of training. However, it can be seen that eventually the critic becomes almost $100\%$ independent after $\approx$8000 episodes.

The experiments show that after a sufficient number of training steps, the learned policies all can drop the external controllers and are efficient to navigate the robot around the obstacles. This verifies that the external controllers have little impact on the final performance of the AsDDPG if the networks are trained with sufficient episodes.


\subsection{Training with Complex Environment and Sparse Reward}

To further validate AsDDPG can learn a good policy by leveraging an external controller, a more complex environment as shown in Fig.\ref{fig:stage complex} is applied together with a sparse reward function. The dense reward function used in the previous experiments alleviates the difficulty of training by leading the robot to decrease its distance to the target for a higher instant reward. But, at the same time, it induces the network to learn a suboptimal policy w.r.t. time because it is also driven by something else besides reaching the target fastest. In this experiment, we use a reward function where the dense part is simply a constant time penalty. This is a challenging reward function for random exploration, as the robot only receives a positive reward when it actually reaches the target. 



\begin{figure}
  \centering
  \includegraphics[width=.8\linewidth]{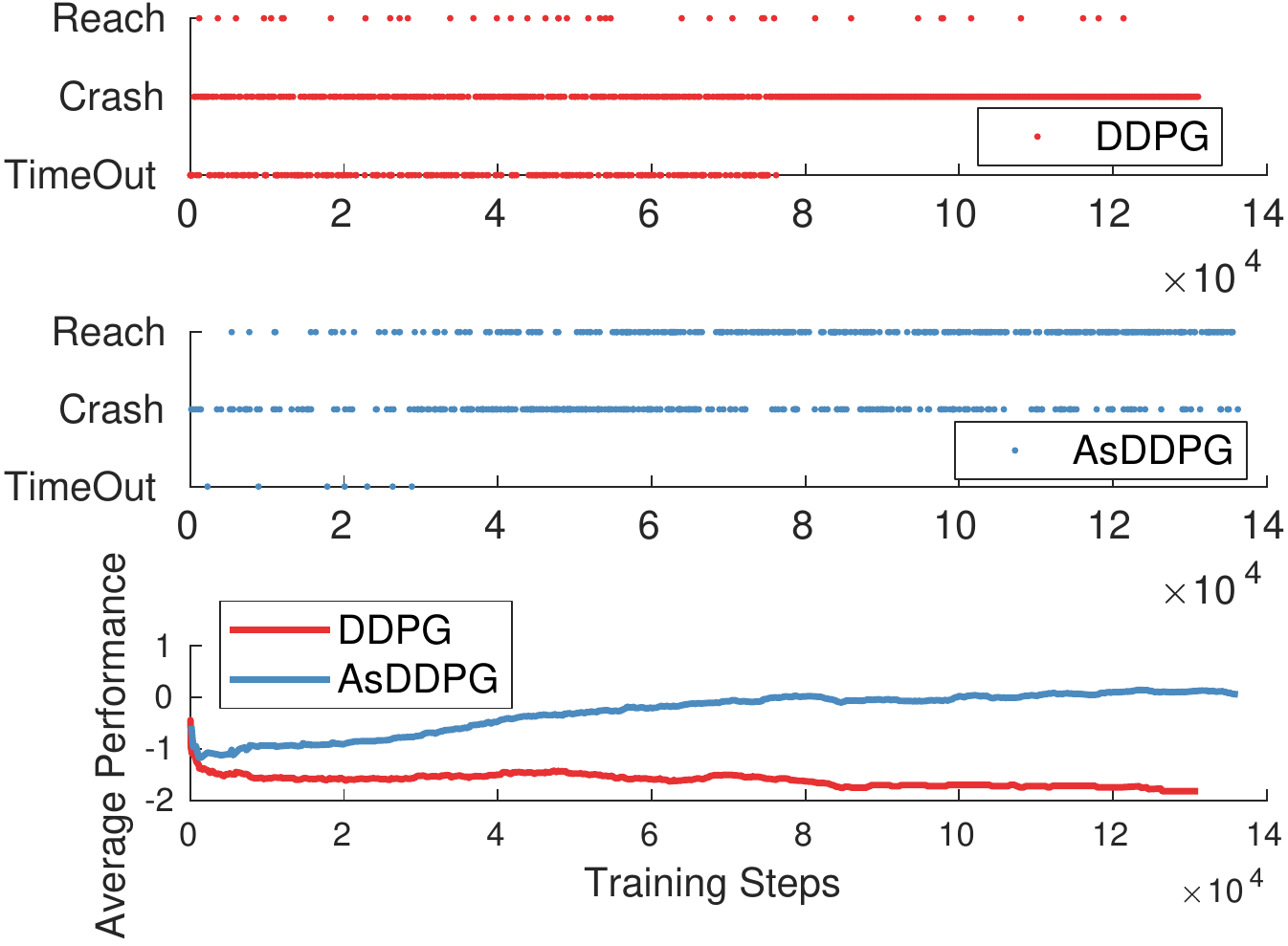}
  \caption{\small The final result (Reach the target, Crash and Time out) for each episode and the smoothed learning curves.\normalsize}
  \label{fig:results for exp b}
\end{figure}

The performances of DDPG and AsDDPG are given in Fig.~\ref{fig:results for exp b}. It can be seen that DDPG seldom learns a proper policy to reach the goal and always collides with the obstacles. More specifically, in the early stages, DDPG runs out of time frequently, which is the worst case due to the accumulative time penalty. After approximately 80k training steps, it learns to crash to avoid the time penalty instead of reaching the goal. In contrast, although AsDDPG also runs out of time at the beginning, it transits to crashing as a better strategy within only 3k steps, and eventually learns to reach the goal for the maximal total reward. This further verifies that the proposed AsDDPG can effectively speed up the training and achieve a better performance with the assistance from the external controller, even in complicated environments with an extremely sparse reward function for which a random exploration guided network can hardly learn a good policy.

\begin{figure}
  \centering
  \includegraphics[width=0.4\linewidth]{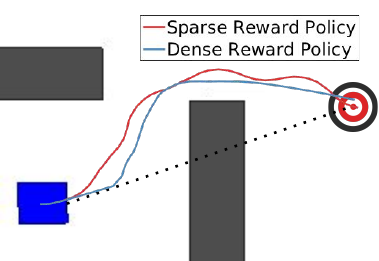}
  \caption{\small Policies learned with dense and sparse reward functions where sparse reward policy takes 43 steps (8.6 sec.) to reach the goal while the dense reward policy takes 53 steps (10.6 sec.).\normalsize}
  \label{fig:traj with different reward function}
\end{figure}

In addition, by using the sparse reward function, the network learns to drive the robot faster and keep a reasonable safe distance to the obstacles. This is shown in Fig.~\ref{fig:traj with different reward function}. With the dense reward, the robot tend to smoothly skirt around the obstacle by slowing down and executing a gentle rotation. However, with a sparse reward, the network learns to plan earlier to avoid obstacles, giving them a wider berth. As such, the robot can travel at a maximum speed, even if the path to the target is not the shortest.

\subsection{Real World Tests}
In the real world experiment, a Pioneer robot which is equipped with a Hokuyo laser scanner is utilized. To localize the robot based on an existing map, we apply the AMCL ROS package. Since our network only acts as a local planner for reaching a near goal without any collision, it is combined with a global planner to achieve a complete navigation system. More specifically, after receiving the destination, the global planner generates a path to the target and each point of the path is transfered into robot's local frame as a network input, together with the laser scans and the speed of the robot.

\begin{figure}
    \centering
    \begin{minipage}{0.42\columnwidth}
        \centering
        \includegraphics[width=\textwidth]{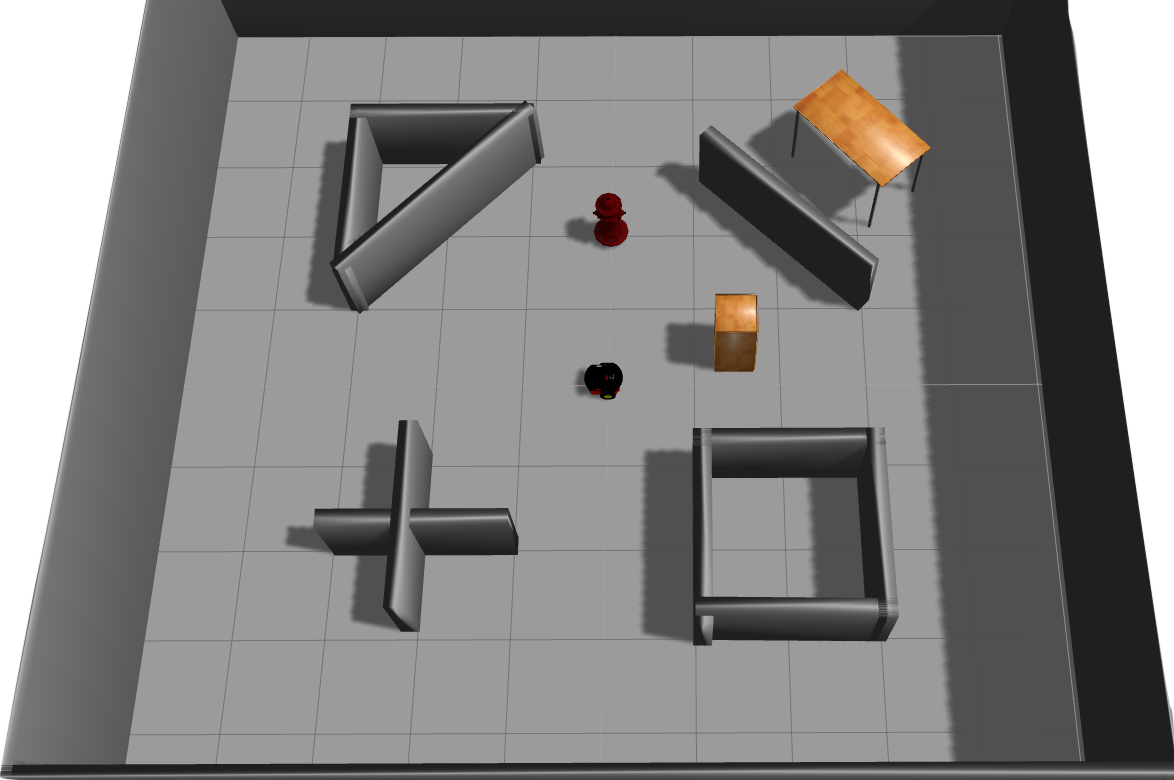} 
        \caption{Gazebo}
        \label{fig:gazebo}
    \end{minipage}\hfill
    \begin{minipage}{0.5\columnwidth}
        \centering
        \includegraphics[width=\textwidth]{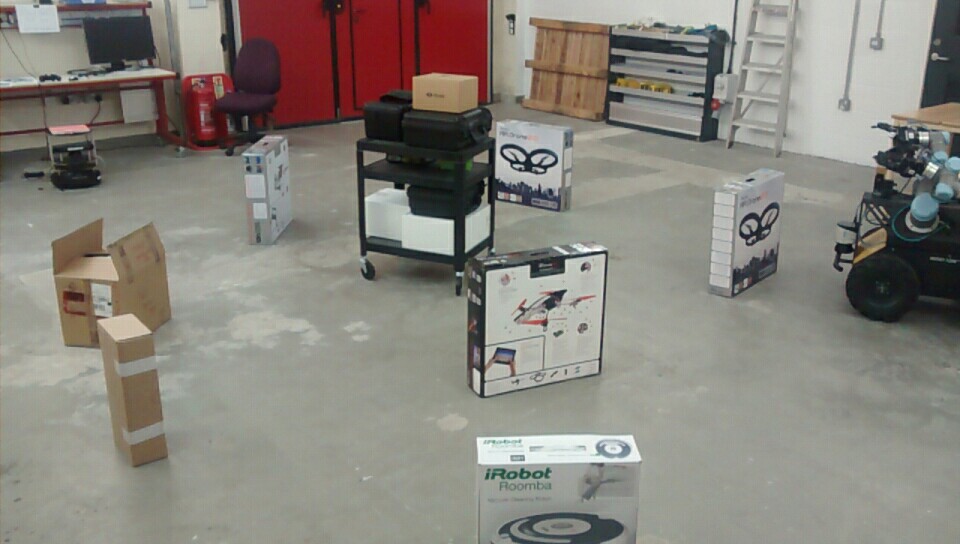} 
        \caption{Real world}
        \label{fig:real world}
    \end{minipage}
\end{figure}

\begin{figure}[t]
  \centering
  \includegraphics[width=.5\linewidth]{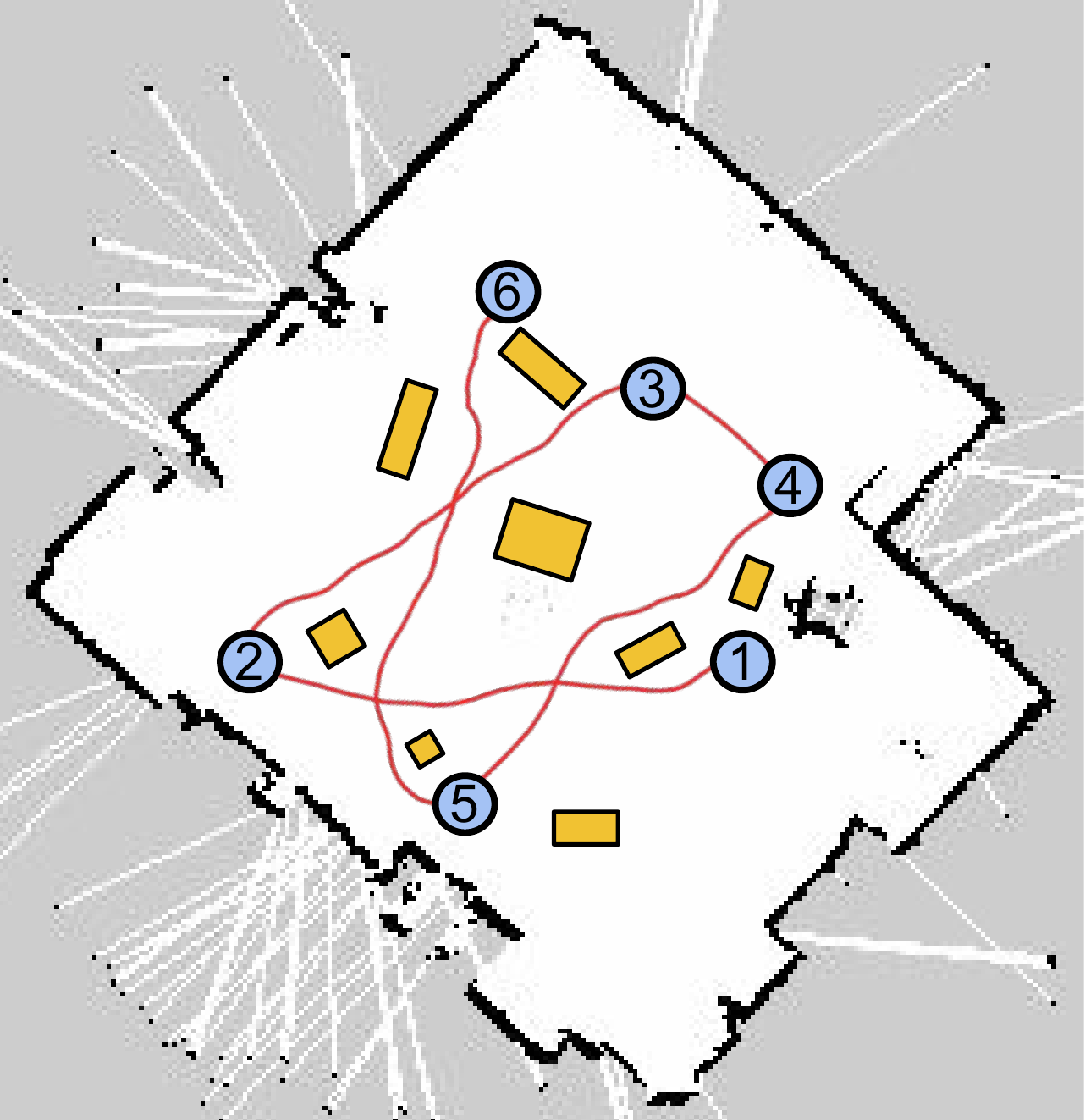}
  \caption{\small The trajectory of robot in the real world experiment. The yellow rectangles are obstacles and blue circles indicate the sequential targets.\normalsize}
  \label{fig:real_traj}
\end{figure}

The simulated Gazebo environment shown in Fig.~\ref{fig:gazebo} is used to traina network. Then, it is directly tested on the real robot in the real world scenario with several obstacles (Fig.~\ref{fig:real world}). In this experiment, a map without any obstacles in the room is established and the robot is driven by the learned policy to reach several target points successively with obstacle avoidance. The robot trajectory and obstacles overlaid on the map are illustrated in Fig.~\ref{fig:real_traj}. The trajectory of the robot is plotted as the red curves which can infer that the robot can smoothly avoid all the obstacles and reach each target successfully.

\section{Conclusions}\label{sec:Conclusion}
 
In this paper, a novel algorithm named Assisted Deep Deterministic Policy Gradient is proposed to tightly combine Deep Reinforcement Learning with an existing controller, achieving faster and more stable learning performance. It harnesses the advantages of both Deep Deterministic Policy Gradient and Deep Q Network. The extensive experiments verify that it can accelerate and effectively stabilize the training procedure for the network in the application of robot navigation even with different hyper-parameters. Furthermore, it can even enable the network to efficiently learn a good policy for some challenging tasks, e.g., navigation in a  complex environment by only using a sparse reward. Real-world experiment also demonstrates the effectiveness of the policy learned by the network.

In the future, the algorithm will be applied in many other scenarios such as robot arm manipulation.

\addtolength{\textheight}{-12cm}   
                                  
\section*{ACKNOWLEDGMENT}
This work was supported in part by EPSRC Robotics and Artificial Intelligence ORCA Hub (grant No. EP/R026173/1) and  EPSRC Mobile Robotics: Enabling a Pervasive Technology of the Future (grant No. EP/M019918/1).

\bibliographystyle{IEEEtran}
\bibliography{ieeeconf}

\begin{thebibliography}{10}
\providecommand{\url}[1]{#1}
\csname url@samestyle\endcsname
\providecommand{\newblock}{\relax}
\providecommand{\bibinfo}[2]{#2}
\providecommand{\BIBentrySTDinterwordspacing}{\spaceskip=0pt\relax}
\providecommand{\BIBentryALTinterwordstretchfactor}{4}
\providecommand{\BIBentryALTinterwordspacing}{\spaceskip=\fontdimen2\font plus
\BIBentryALTinterwordstretchfactor\fontdimen3\font minus
  \fontdimen4\font\relax}
\providecommand{\BIBforeignlanguage}[2]{{%
\expandafter\ifx\csname l@#1\endcsname\relax
\typeout{** WARNING: IEEEtran.bst: No hyphenation pattern has been}%
\typeout{** loaded for the language `#1'. Using the pattern for}%
\typeout{** the default language instead.}%
\else
\language=\csname l@#1\endcsname
\fi
#2}}
\providecommand{\BIBdecl}{\relax}
\BIBdecl

\bibitem{mnih2015human}
V.~Mnih, K.~Kavukcuoglu, D.~Silver, A.~A. Rusu, J.~Veness, M.~G. Bellemare,
  A.~Graves, M.~Riedmiller, A.~K. Fidjeland, G.~Ostrovski \emph{et~al.},
  ``Human-level control through deep reinforcement learning,'' \emph{Nature},
  vol. 518, no. 7540, pp. 529--533, 2015.

\bibitem{mnih2016asynchronous}
V.~Mnih, A.~P. Badia, M.~Mirza, A.~Graves, T.~Lillicrap, T.~Harley, D.~Silver,
  and K.~Kavukcuoglu, ``Asynchronous methods for deep reinforcement learning,''
  in \emph{ICML}, 2016, pp. 1928--1937.

\bibitem{schaul2015prioritized}
T.~Schaul, J.~Quan, I.~Antonoglou, and D.~Silver, ``Prioritized experience
  replay,'' \emph{arXiv preprint arXiv:1511.05952}, 2015.

\bibitem{tai2017virtual}
L.~Tai, G.~Paolo, and M.~Liu, ``Virtual-to-real deep reinforcement learning:
  Continuous control of mobile robots for mapless navigation,'' \emph{IROS},
  2017.

\bibitem{xie2017towards}
L.~Xie, S.~Wang, A.~Markham, and N.~Trigoni, ``Towards monocular vision based
  obstacle avoidance through deep reinforcement learning,'' \emph{arXiv
  preprint arXiv:1706.09829}, 2017.

\bibitem{pfeiffer2017perception}
M.~Pfeiffer, M.~Schaeuble, J.~Nieto, R.~Siegwart, and C.~Cadena, ``From
  perception to decision: A data-driven approach to end-to-end motion planning
  for autonomous ground robots,'' in \emph{ICRA}.\hskip 1em plus 0.5em minus
  0.4em\relax IEEE, 2017, pp. 1527--1533.

\bibitem{yang2017obstacle}
S.~Yang, S.~Konam, C.~Ma, S.~Rosenthal, M.~Veloso, and S.~Scherer, ``Obstacle
  avoidance through deep networks based intermediate perception,'' \emph{arXiv
  preprint arXiv:1704.08759}, 2017.

\bibitem{duan2017one}
Y.~Duan, M.~Andrychowicz, B.~Stadie, J.~Ho, J.~Schneider, I.~Sutskever,
  P.~Abbeel, and W.~Zaremba, ``One-shot imitation learning,'' \emph{arXiv
  preprint arXiv:1703.07326}, 2017.

\bibitem{lillicrap2015continuous}
T.~P. Lillicrap, J.~J. Hunt, A.~Pritzel, N.~Heess, T.~Erez, Y.~Tassa,
  D.~Silver, and D.~Wierstra, ``Continuous control with deep reinforcement
  learning,'' \emph{ICLR}, 2016.

\bibitem{wang2017deepvo}
S.~Wang, R.~Clark, H.~Wen, and N.~Trigoni, ``Deepvo: towards end-to-end visual
  odometry with deep recurrent convolutional neural networks,'' in
  \emph{ICRA}.\hskip 1em plus 0.5em minus 0.4em\relax IEEE, 2017, pp.
  2043--2050.

\bibitem{Clark_AAAI17}
R.~Clark, S.~Wang, H.~Wen, A.~Markham, and N.~Trigoni, ``Vinet: Visual-inertial
  odometry as a sequence-to-sequence learning problem,'' in \emph{AAAI}, 2017,
  pp. 3995--4001.

\bibitem{levine2016learning}
S.~Levine, P.~Pastor, A.~Krizhevsky, J.~Ibarz, and D.~Quillen, ``Learning
  hand-eye coordination for robotic grasping with deep learning and large-scale
  data collection,'' \emph{IJRR}, p. 0278364917710318, 2016.

\bibitem{oriolo1995line}
G.~Oriolo, M.~Vendittelli, and G.~Ulivi, ``On-line map building and navigation
  for autonomous mobile robots,'' in \emph{Robotics and Automation, 1995.
  Proceedings., 1995 IEEE International Conference on}, vol.~3.\hskip 1em plus
  0.5em minus 0.4em\relax IEEE, 1995, pp. 2900--2906.

\bibitem{kim1999symbolic}
D.~Kim and R.~Nevatia, ``Symbolic navigation with a generic map,''
  \emph{Autonomous Robots}, vol.~6, no.~1, pp. 69--88, 1999.

\bibitem{giusti2016machine}
A.~Giusti, J.~Guzzi, D.~C. Cire{\c{s}}an, F.-L. He, J.~P. Rodr{\'\i}guez,
  F.~Fontana, M.~Faessler, C.~Forster, J.~Schmidhuber, G.~Di~Caro
  \emph{et~al.}, ``A machine learning approach to visual perception of forest
  trails for mobile robots,'' \emph{RA Letters}, vol.~1, no.~2, pp. 661--667,
  2016.

\bibitem{pfeiffer2016perception}
M.~Pfeiffer, M.~Schaeuble, J.~Nieto, R.~Siegwart, and C.~Cadena, ``From
  perception to decision: A data-driven approach to end-to-end motion planning
  for autonomous ground robots,'' \emph{arXiv:1609.07910}, 2016.

\bibitem{gandhi2017learning}
D.~Gandhi, L.~Pinto, and A.~Gupta, ``Learning to fly by crashing,'' \emph{arXiv
  preprint arXiv:1704.05588}, 2017.

\bibitem{sadeghi2016cad}
F.~Sadeghi and S.~Levine, ``(cad)2rl: Real single-image flight without a single
  real image,'' \emph{Robotics: Science and Systems}, 2017.

\bibitem{zhang2016deep}
J.~Zhang, J.~T. Springenberg, J.~Boedecker, and W.~Burgard, ``Deep
  reinforcement learning with successor features for navigation across similar
  environments,'' \emph{arXiv preprint arXiv:1612.05533}, 2016.

\bibitem{gu2016continuous}
S.~Gu, T.~Lillicrap, I.~Sutskever, and S.~Levine, ``Continuous deep q-learning
  with model-based acceleration,'' in \emph{ICML}, 2016, pp. 2829--2838.

\bibitem{zhang2016learning}
T.~Zhang, G.~Kahn, S.~Levine, and P.~Abbeel, ``Learning deep control policies
  for autonomous aerial vehicles with mpc-guided policy search,'' in
  \emph{ICRA}.\hskip 1em plus 0.5em minus 0.4em\relax IEEE, 2016, pp. 528--535.

\bibitem{vevcerik2017leveraging}
M.~Ve{\v{c}}er{\'\i}k, T.~Hester, J.~Scholz, F.~Wang, O.~Pietquin, B.~Piot,
  N.~Heess, T.~Roth{\"o}rl, T.~Lampe, and M.~Riedmiller, ``Leveraging
  demonstrations for deep reinforcement learning on robotics problems with
  sparse rewards,'' \emph{arXiv preprint arXiv:1707.08817}, 2017.

\bibitem{tessler2017deep}
C.~Tessler, S.~Givony, T.~Zahavy, D.~J. Mankowitz, and S.~Mannor, ``A deep
  hierarchical approach to lifelong learning in minecraft.'' in \emph{AAAI},
  vol.~3, 2017, p.~6.

\bibitem{DBLP:journals/corr/WangFL15}
Z.~Wang, N.~de~Freitas, and M.~Lanctot, ``Dueling network architectures for
  deep reinforcement learning,'' \emph{CoRR}, vol. abs/1511.06581, 2015.

\bibitem{van2016deep}
H.~Van~Hasselt, A.~Guez, and D.~Silver, ``Deep reinforcement learning with
  double q-learning.'' in \emph{AAAI}, 2016, pp. 2094--2100.

\bibitem{sutton2000policy}
R.~S. Sutton, D.~A. McAllester, S.~P. Singh, and Y.~Mansour, ``Policy gradient
  methods for reinforcement learning with function approximation,'' in
  \emph{NIPS}, 2000, pp. 1057--1063.

\end{thebibliography}

\end{document}